\documentclass[11pt]{article}

\usepackage[preprint]{acl}

\usepackage{times}
\usepackage{latexsym}
\usepackage{booktabs}  
\usepackage{graphicx}
\usepackage{amsmath}
\usepackage{enumitem}
\usepackage{url}
\usepackage{amssymb}
\usepackage{multirow}
\usepackage{tabularx}
\usepackage[table]{xcolor}
\usepackage{colortbl}
\usepackage{mdframed}
\usepackage{xcolor}
\usepackage{caption}

\usepackage{graphicx}
\usepackage[T1]{fontenc}

\usepackage[utf8]{inputenc}

\usepackage{microtype}

\usepackage{inconsolata}

\usepackage{graphicx}

\usepackage{multirow}
\usepackage{pifont}
\usepackage{siunitx}

\newcommand{\bench}{LR-Bench}
\newcolumntype{g}{>{\columncolor[HTML]{EAF0F6}}c}

\title{RATE: Reviewer Profiling and Annotation-free Training for Expertise Ranking in Peer Review Systems}

\author{
  Weicong Liu\thanks{Equal contribution.} \quad
  Zixuan Yang\footnotemark[1] \quad
  Yibo Zhao \quad
  Xiang Li\thanks{Corresponding Author: \texttt{xiangli@dase.ecnu.edu.cn}} \\
  School of Data Science and Engineering, East China Normal University
}

\begin{document}
\maketitle
\begin{abstract}
Reviewer assignment is increasingly critical yet challenging in the LLM era, where rapid topic shifts render many pre-2023 benchmarks outdated and where proxy signals poorly reflect true reviewer familiarity. We address this evaluation bottleneck by introducing LR-bench, a high-fidelity, up-to-date benchmark curated from 2024--2025 AI/NLP manuscripts with five-level self-assessed familiarity ratings collected via a large-scale email survey, yielding 1{,}055 expert-annotated paper--reviewer--score annotations. 
We further propose RATE, a reviewer-centric ranking framework that distills each reviewer’s recent publications into compact keyword-based profiles and fine-tunes an embedding model with weak preference supervision constructed from heuristic retrieval signals, enabling the matching of each manuscript against a reviewer profile directly.
Across the LR-bench and the CMU gold-standard dataset, our approach consistently achieves state-of-the-art performance, outperforming strong embedding baselines by a clear margin. We release LR-bench at \url{https://huggingface.co/datasets/Gnociew/LR-bench}, and an github repository at \url{https://github.com/Gnociew/RATE-Reviewer-Assignment}.
\end{abstract}

\section{Introduction}

As a cornerstone of modern scientific research, the peer review system plays a crucial role in helping scientists evaluate submissions, providing constructive feedback, and safeguarding academic integrity~\cite{Black1998,Thurner_2011,10.5555/2888619.2889159}. An expert reviewer can substantially improve a manuscript; however, if a reviewer lacks relevant domain expertise, the process may waste time for both authors and reviewers. Consequently, the selection of reviewers requires significant care. With the rapid growth of computer science, especially in the area of artificial intelligence, the number of submissions to conferences and journals has surged~\cite{doi:10.1177/01655515231176668,Shah2022AnOO}, making manual reviewer assignment increasingly impractical. Therefore, developing effective reviewer-assignment algorithms has become essential.

\begin{figure}[t]
  \centering
  \includegraphics[width=\linewidth]{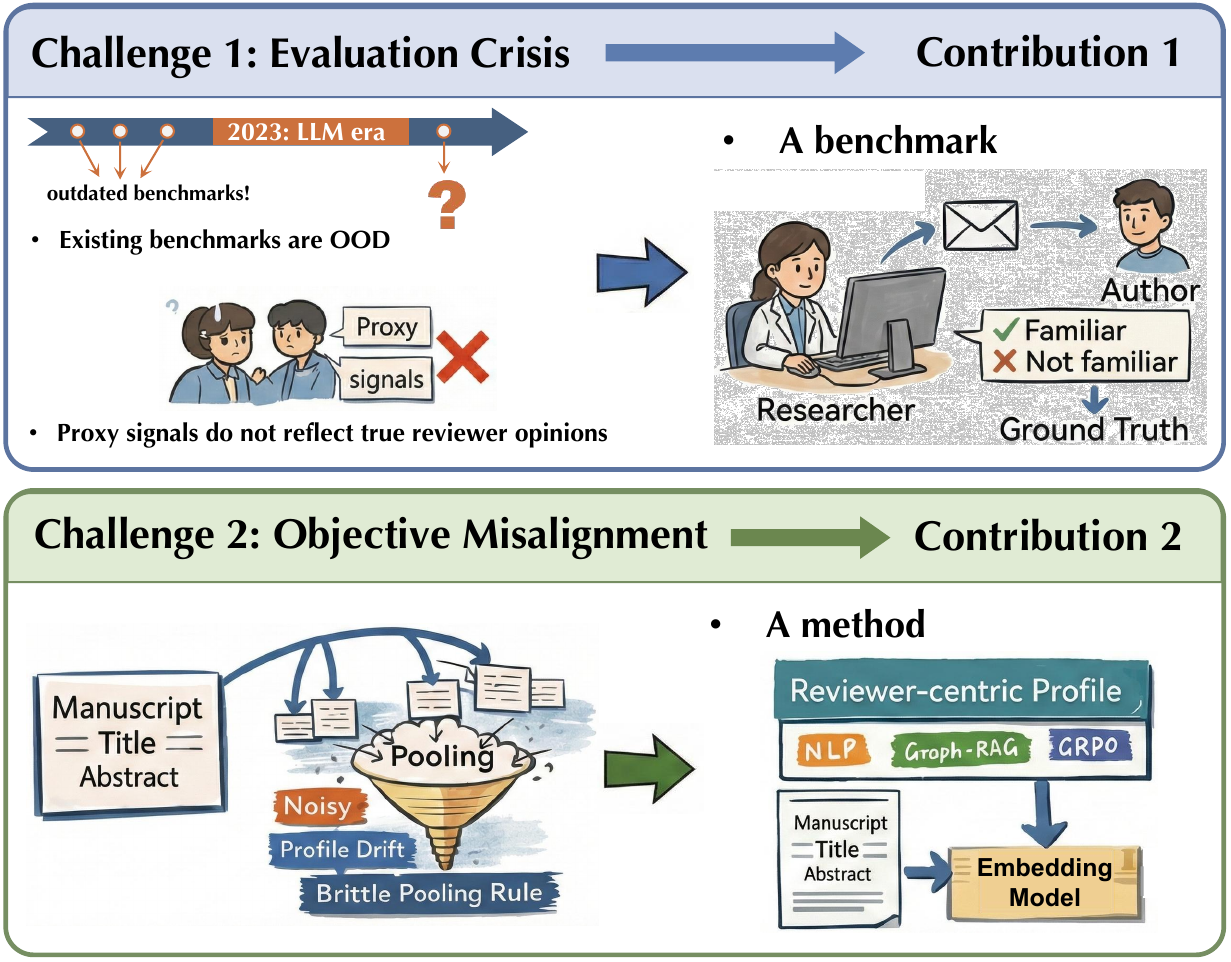}
  \caption{Challenges of reviewer assignment.}
  \label{fig:intro}
\end{figure}


Despite its significance, the field of reviewer assignment currently faces two critical challenges. 
As illustrated in Figure~\ref{fig:intro}, the first challenge lies in the evaluation crisis: \textbf{the lack of high-fidelity, open-source, and up-to-date benchmarks.}
The exponential surge in AI and NLP research, particularly in the Large Language Model (LLM) era, has created a severe temporal shift. Existing benchmarks~\cite{CMU,SciRepEval_dataset}, most of which were created before 2023, are increasingly out-of-distribution (OOD) and fail to reflect contemporary research topics. Further, many datasets~\cite{NIPS_dataset, CoF} rely on third-party annotations rather than direct expert feedback, which often fail to capture the nuanced expertise.

{The second challenge is the \textbf{misalignment between training objectives and the inference goal} in prior work. Many methods~\cite{CMU,CoF,Hsieh2024VulnerabilityOT} are optimized to retrieve papers that are most similar to the target manuscript, rather than to retrieve reviewers who are truly suitable for it. As a result, they do not directly model a reviewer’s expertise; instead, they approximate reviewer relevance by aggregating similarities between the manuscript and the reviewer’s publications using simple pooling rules (e.g., mean, max, or percentile). This design is sensitive to noise in a reviewer’s publication list, leading to ``profile drift'', and it also relies on manually choosing a pooling strategy, which is often brittle and dataset-dependent.
For example, a reviewer whose primary expertise is retrieval-augmented generation (RAG) may have coauthored a paper on graph learning without being a domain expert in graphs. Under max pooling, this single off-topic publication can dominate the aggregated score, causing the reviewer to be incorrectly ranked highly for a graph-focused manuscript.}


To address the lack of high-fidelity, up-to-date benchmarks, we first introduce \bench, a high-fidelity and up-to-date benchmark specifically designed to reflect the contemporary research landscape.
Our benchmark centers on manuscripts curated from leading AI and NLP conferences within the last two years (2024–2025), directly bridging the content gap created by the recent surge in LLM research.
To ensure high data quality, we employ a multi-stage curation process: for each manuscript, we retrieve a candidate pool of reviewers via content-based filtering, and subsequently collect five-level self-assessed Likert familiarity ratings through a large-scale email survey.
This approach yields 1,055 high-fidelity paper-reviewer-score pairs, providing a gold standard reflecting real-world expert judgment.

Besides the benchmark, we propose a novel reviewer profiling-ranking algorithm, RATE, which moves beyond the limitations of heuristic pooling.
To overcome the limitations of heuristic pooling and profile drift, we propose a novel LLM-augmented reviewer profiling and self-supervised ranking framework.
Specifically, we utilize LLMs to distill core keywords from a reviewer’s publication history, synthesizing them into a structured natural language profile that captures their essential expertise.
To train a robust assignment model without manual labels, we develop an automated data construction scheme based on pseudo-labeling.
For any target manuscript, we retrieve potential candidates via semantic similarity and employ BM25\cite{10.1561/1500000019,dual,fensore2025evaluatinghybridretrievalaugmented} scores between the manuscript and reviewer profiles as a weak supervision signal to identify positive and hard negative pairs.
Through contrastive learning on these synthesized pairs, our model learns to bridge the gap between explicit keyword matching and deep semantic expertise alignment.
Experimental results on both \bench~and the CMU gold standard dataset~\cite{CMU} demonstrate that our approach achieves state-of-the-art performance, even surpassing many existing methods that rely on expensive human-annotated data.

In summary, our contributions are as follows:

\begin{itemize}[leftmargin=*, itemsep=0pt]
\item \textbf{A high-fidelity contemporary benchmark}: We release \bench, including 1,055 paper-reviewer-score pairs from 2024-2025, establishing a high-fidelity ground truth for the rigorous evaluation of modern assignment systems.
\item \textbf{An LLM-based profiling method}: We synthesize reviewer expertise into structured natural language profiles using LLM-distilled keywords, effectively eliminating the profile drift issues in traditional heuristic approaches.
\item \textbf{Zero-annotation training paradigm}: We design a self-supervised training strategy using BM25-guided pseudo-labeling, which allows the model to learn deep semantic expertise alignment without the need for any human-annotation.
\end{itemize}

\section{Related Work}

\subsection{Reviewer Assignment Benchmarks}


Existing benchmarks for reviewer assignment are primarily divided into two categories based on their annotation sources. The first category leverages proxy signals, such as authorship and keyword similarity~\cite{OpenReview, SIGIR_dataset,SciRepEval_dataset}, or manual labels provided by third-party annotators~\cite{NIPS_dataset, CoF}. While these approaches are more accessible, they often introduce significant noise and suffer from limited annotation fidelity. The second category relies on reviewer self-assessments~\cite{10.1145/133160.133205, 10.1145/1458082.1458127, CMU}, which are widely regarded as the ``gold standard'' due to their high reliability. However, these datasets are difficult to collect and many existing ones have become outdated, failing to capture recent research trends. In this work, we follow the second paradigm and introduce a contemporary dataset and construction pipeline that continuously scales while preserving high-fidelity, up-to-date relevance labels.

\subsection{Reviewer Assignment Methods}

{Prior work on reviewer assignment is typically framed as a two-stage process: {estimating paper–reviewer relevance} and {allocating reviewers to papers} under practical constraints. In the relevance estimation stage, existing methods can be broadly categorized into three classes: (i) explicit-feedback approaches~\cite{https://doi.org/10.1002/int.1055,Tayal2014,pmlr-v124-fiez20a} that leverage human-provided signals such as reviewer interests and bids; (ii) content-based approaches~\cite{Tan2021,aitymbetov-zorbas-2025-autonomous,CoF} that compute similarity between reviewer profiles and the submitted manuscript using lexical matching, statistical models, or embedding-based representations; and (iii) network-based approaches~\cite{10.1145/1458082.1458127,6961199} that exploit relational signals from citation, co-authorship, or collaboration graphs. These approaches are often complementary and are frequently integrated into multi-stage retrieval-and-reranking pipelines.}
Following relevance estimation, the allocation stage focuses on optimizing assignments to satisfy various objectives such as aggregate similarity \cite{5d21c3eea4d146f6a345dd044b4ff829,TPMS}, topic coverage \cite{10.1145/1645953.1646207}, or fairness \cite{Garg2010,Stelmakh2018PeerReview4AllFA}. These frameworks generally model the problem as a constrained optimization task to limit reviewer workload and meet review requirements. In this work, we focus on the stage of estimating paper-reviewer relevance and introduce a novel algorithm that provides more accurate scores for matching.

\begin{table*}[t]
\centering
\small
\setlength{\tabcolsep}{5pt}
\renewcommand{\arraystretch}{1.08}
\caption{Benchmark statistics characterizing coverage, supervision density, and rating distribution.}
\label{tab:benchmark_stats}

\begin{tabular}{ccccccccc}
\toprule
\multirow{2}{*}{\textbf{Item}} & \textbf{\multirow{2}{*}{\raisebox{-0.7\height}{\shortstack[c]{Total\\ Counts}}}}
 & \multirow{2}{*}{\textbf{Metric}} & \multicolumn{6}{c}{\textbf{Value}} \\
\cmidrule(lr){4-9}
 & &  & 1 & 2 & 3 & 4 & 5 & 6 \\
\midrule

\multirow{2}{*}{\shortstack[c]{Query Papers\\ per Annotator}} & \multirow{2}{*}{406} 
& Count & 196 & 61  & 31  & 21  & 22  & 75  \\
& & \%    & 48.28 & 15.02 & 7.64 & 5.17 & 5.42 & 18.47 \\
\midrule

\multirow{2}{*}{\shortstack[c]{Annotators per\\ Query Paper}} & \multirow{2}{*}{825 }
& Count & 640 & 150 & 27  & 6   & 2   & 0   \\
& & \%    & 77.58 & 18.18 & 3.27 & 0.73 & 0.24 & 0.00 \\
\midrule

\multirow{2}{*}{\shortstack[c]{Rating\\ Distribution}} & \multirow{2}{*}{1055 }
& Count & 100 & 201 & 305 & 273 & 176 & -- \\
& & \%    & 9.48 & 19.05 & 28.91 & 25.88 & 16.68 & -- \\
\bottomrule
\end{tabular}
\end{table*}

\section{Dataset Construction}\label{sec:dataset}

To address the challenge of lacking high-fidelity, open-source, and up-to-date benchmarks, we propose \bench, which consists of (i) a large unlabeled arXiv corpus used to construct paper/author metadata and candidate reviewer pools, and (ii) a labeled benchmark subset with self-reported expertise ratings collected via email outreach.

\subsection{Data Source \& Preprocessing}\label{sec:arxiv}

{We construct an up-to-date, unlabeled corpus by crawling recent papers from arXiv, the largest publicly accessible and continuously updated preprint repository with broad coverage across computer science. We focus on five representative sub-areas: Artificial Intelligence (\texttt{cs.AI}), Computation and Language (\texttt{cs.CL}), Computer Vision and Pattern Recognition (\texttt{cs.CV}), Information Retrieval (\texttt{cs.IR}), and Machine Learning (\texttt{cs.LG}). Collectively, these sub-areas cover major CS/AI research areas with broad topical diversity and abundant recent submissions, providing a practical testbed for reviewer assignment in modern CS venues. To stay aligned with fast-evolving research trends in the LLM era and to reduce temporal drift in topics and expertise signals, we restrict the collection to papers whose last revision date falls within a two-year window (Oct 2023-Oct 2025).

{For each paper, we retrieve the external metadata, including the title, arXiv identifier, and PDF URL via the arXiv API. Subsequently, we download the corresponding PDFs and employ GROBID~\citep{lopez2009grobid} to extract the title, abstract, author list, affiliations (when available) and email addresses (when available). To ensure data integrity, we perform a cross-source consistency check: a paper is discarded if its PDF-extracted title deviates significantly from the arXiv metadata. Furthermore, we apply rigorous filtering to exclude entries with missing essential fields (title, abstract, or authors) or corrupted text. For papers with multiple revisions, only the most recent version within our temporal window is retained. The corpus comprises 161,228 unique papers with high-fidelity metadata.}

{To ensure consistent author identities for downstream reviewer assignment, we perform author disambiguation via a precision-oriented hierarchical strategy. 
Following metadata normalization, we reconcile author entries by prioritizing exact email matches, followed by exact affiliation matches when emails are unavailable. For non-exact matches (e.g., institution variants or abbreviations), we employ an LLM to semantically verify identity. 
Pairs are merged only upon exact metadata alignment or LLM confirmation; otherwise, they remain distinct. 
While this conservative approach may leave some cross-email identities split, it effectively mitigates homonym conflation under sparse metadata. The resulting corpus identifies 513,877 unique authors, providing a robust foundation for paper-author relations.}

\subsection{Query Sampling \& Candidate Recall}\label{sec:recall}

Based on the full unlabeled corpus, we sample 4,000 query papers for ground-truth collection using equal-size stratified sampling across the five sub-fields (800 papers per sub-area).
For each paper $p$ in our corpus, we formulate its textual representation $x_p$ by concatenating its title and abstract. We then employ a pre-trained Sentence-BERT encoder to map each $x_p$ into a dense embedding space. All paper embeddings are indexed in a high-performance vector database faiss~\cite{faiss} to facilitate efficient search.

To construct a recall set of potential reviewers for each query paper, we implement a two-stage pipeline consisting of optimistic dense vector retrieval and Conflict of Interest (COI)~\cite{5616179,10.1145/3183713.3193552,10.1016/j.artint.2024.104119} filtering.
To identify potential reviewers for a given query paper $q$, we retrieve all papers from the database with a similarity score exceeding a predefined threshold $\tau$. The initial candidate reviewer set, denoted as $\mathcal C_q$, is formed by the authors of these retrieved papers. To ensure the integrity and objectivity of the assignment process, we implement a rigorous COI filtering mechanism. Specifically, we exclude (i) any author of the query manuscript itself and (ii) any individual who shares a direct co-authorship history with the authors of $q$. This procedure yields the final refined candidate reviewer set $\mathcal C_q^\text{COI}$ for the subsequent ground-truth collection.

\begin{figure*}[t]
  \centering
  \includegraphics[width=\textwidth]{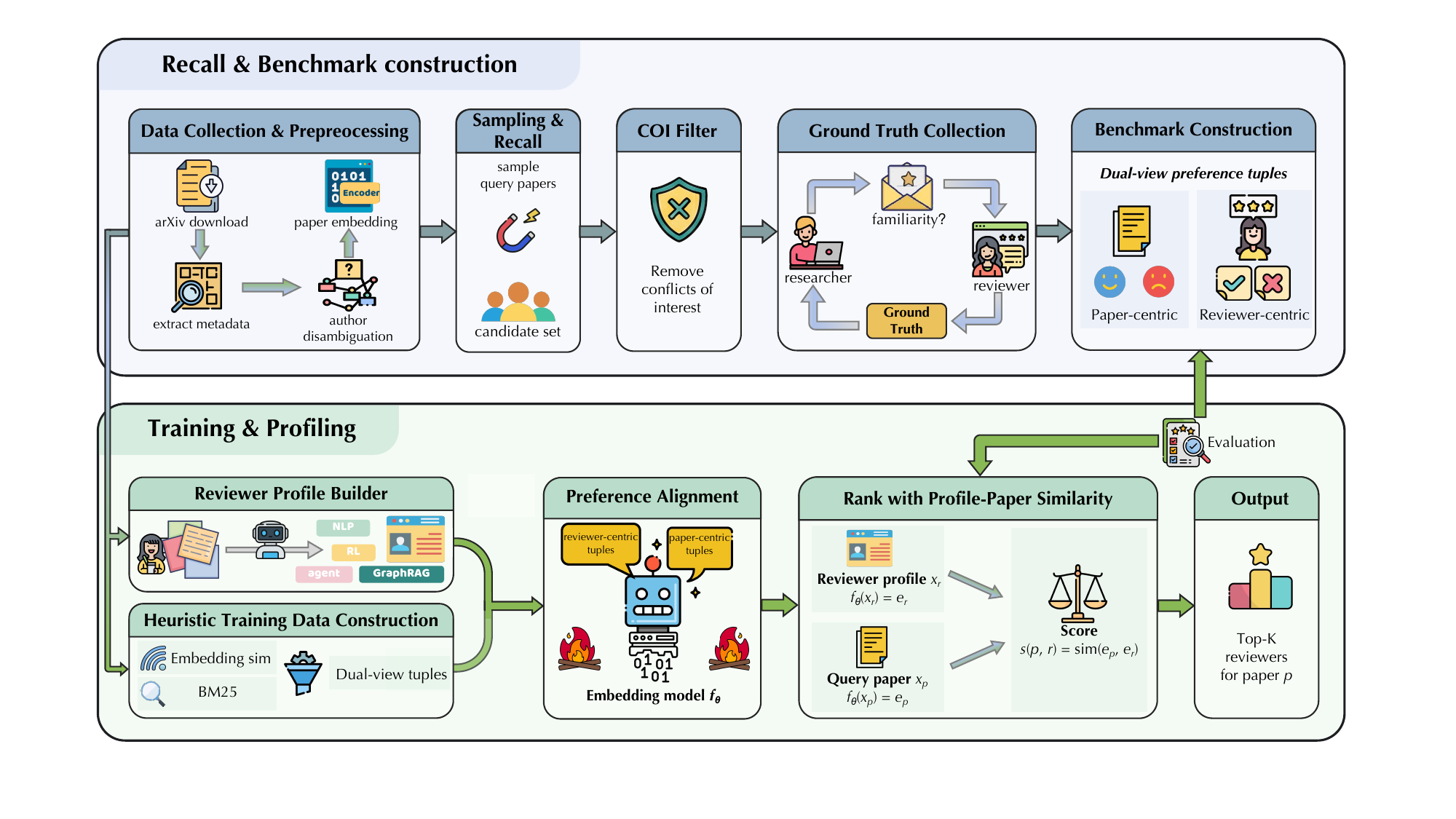}
  \caption{Overview of pipline.}
  \label{fig:framework}
\end{figure*}



\subsection{Expertise Ground Truth Collection}


We collect ground-truth via email outreach: we contact candidates in the reviewer pool and ask them to rate their familiarity/expertise for the query papers they are matched to by our retrieval pipeline, after reading the title and abstract of each paper. This aims to capture reviewer expertise directly rather than relying on weak proxies such as topical overlap. To help reduce subjectivity and encourage consistent interpretation across participants, we adopt a five-level Behaviorally Anchored Rating Scale (BARS) ~\cite{Garine2014TheCA,https://doi.org/10.1002/ets2.12152} ranging from top expert to no expertise. Details of the scale are provided in Appendix~\ref{subsec:expert_scale}.

To minimize disruption to candidate reviewers, we aggregate and consolidate assignments for each candidate across queries and cap the workload at six query papers per contacted candidate. Each outreach email explicitly states the purpose and intended use of the collected feedback; the full email template is included in the Appendix~\ref{sec:appendix_email}. By December 19, 2025, we received 407 responses, yielding 1,069 reviewer–paper ratings. We then performed data cleaning and quality control to reduce the impact of low-effort or unreliable responses. First, we removed duplicate ratings for the same reviewer–paper pair, keeping only the most recent submission (8 duplicate ratings removed). Next, we computed summary statistics to identify potentially suspicious patterns, such as near-zero rating variance or highly imbalanced scores that appear insensitive to paper content. This process flagged 23 ratings for further inspection, and manual review led to the removal of 6 additional ratings. Finally, we obtain 1,055 reviewer-paper ratings.



\subsection{Benchmark Construction \& Statistics}\label{sec:benchmark}

Finally, we convert the labeled subset into an evaluation benchmark and report key statistics.

The resulting benchmark consists of pointwise and pairwise supervision derived from the collected ratings. Each raw annotation is a pointwise record $(p,r,y)$, where $p$ is a query paper, $r$ is a candidate reviewer, and $y\in \{1,…,5\}$ is the BARS expertise rating. From these, we derive pairwise preference tuples to support ranking-based training and evaluation: (i) paper-centric tuples $(p,r^+,r^-)$, where $r^+$ is rated as more expert than $r^-$ for the same paper, and (ii) reviewer-centric tuples $(r,p^+,p^-)$, where the reviewer indicates higher familiarity with $p^+$ than $p^-$. In both cases, we form a preference whenever the two ratings differ, treating the higher-rated item as preferred and discarding ties. In total, the benchmark comprises 214 paper-centric and 1,185 reviewer-centric tuples.

We report coverage and supervision density in Table~\ref{tab:benchmark_stats}, including (1) queries per participant (capped at 6), (2) participants per paper, and (3) the $\{1, \dots, 5\}$ rating distribution, exposing the benchmark's sparsity and long-tail behavior.

\section{Reviewer-centric Ranking}

To address the challenge of misalignment between training objectives and the inference goal, we introduce RATE, a pipeline that contains \underline{R}eviewer profiling and \underline{A}nnotation-free \underline{T}raining for \underline{E}xpertise ranking in the peer reviewer assignment system. We formalize this task as a ranking problem, in which the query and candidate items are heterogeneous and may not exhibit aligned or comparable semantics. 
Given a query manuscript $q$ and a candidate reviewer set $\mathcal C_q^\text{COI}$, our goal is to learn a scoring model $f_\theta$. This model computes a relevance score between each candidate reviewer $r \in \mathcal C_q^\text{COI}$ and the query $q$, which is subsequently used to rank the candidates in descending order of predicted score.

\subsection{Reviewer Profiling}\label{sec:profiling}

To better capture a reviewer's expertise and address the profile drift issue, we depart from the traditional approach that computes pairwise similarity between the query manuscript and each of the reviewer's historical publications before aggregating these individual scores. Instead, we leverage the extensive internal knowledge and zero-shot capabilities of LLMs to synthesize the reviewer's entire publication history into a cohesive profile. Then, we can obtain a single, holistic embedding for this synthesized profile to perform the final matching. This ``synthesize-then-embed'' strategy allows the model to grasp a unified research trajectory rather than relying on paper-level comparisons.

Specifically, for each reviewer $r$ and their associated publication history $\mathcal P_r$, we employ an LLM to distill a list of salient keywords from each individual paper $p\in \mathcal P_r$. These keyword lists are then aggregated into a single comprehensive keyword collection. Notably, we deliberately retain duplicate keywords during this aggregation, which ensures that the frequency of recurrence for specific terms serves as a proxy for the reviewer's level of expertise and familiarity within those particular research sub-domains.
Finally, we linearize this frequency-preserving collection into a natural-language sentence by appending the keywords, joined by commas, to a fixed prefix (e.g., ``\emph{The reviewer’s research keywords include:}''). This yields a unified textual representation $x_r$ for the reviewer profile $r$. With this textual profile, reviewer-paper matching is cast as a heterogeneous text retrieval problem, where reviewer expertise descriptions and paper abstracts play distinct semantic roles but are required to be aligned in a shared embedding space.

\subsection{Dual-view Heuristic Preference Alignment}

Despite being framed within a text-based retrieval interface, reviewer–paper matching fundamentally involves heterogeneous and semantically asymmetric text types, which general-purpose embedding models struggle to represent in specialized domains. 
Therefore, we propose a dual-view, annotation-free preference data construction pipeline to adapt these models to this specific task.
To unify the optimization process, we define each training instance as a preference triplet $(a, c^+, c^-)$, where $a$ represents the anchor and $c$ represents a candidate. This formulation covers two symmetric views:
Paper-centric view ($a=q, c=r$): Finding the better-matched reviewer $r^+$ over $r^-$ for a manuscript $q$.
Reviewer-centric view ($a=r, c=q$): Identifying the query paper $q^+$ that better aligns with a reviewer $r$'s expertise than $q^-$.
This dual-perspective optimization helps the model learn a more robust representation of expertise space~\cite{dual,ye2024dual}.


To enable the model to recognize that keyword frequency directly reflects a reviewer’s expertise in a sub-domain, we employ BM25, a lexical retrieval method sensitive to term frequency, to construct our training data. Our core philosophy is to prioritize high precision over high recall; we prefer to exclude potentially noisy hard samples to ensure the cleanliness of the training signal. Specifically, we apply this logic to both paper-centric and reviewer-centric pairs. For any given anchor (either a query paper or a reviewer profile), we rank its corresponding candidates using BM25. We select the top-ranked candidate as the positive sample ($r^+$ or $q^+$). To control difficulty, we choose candidates with scores approximately one-tenth and one-third of the positive score as easy and hard negatives, respectively. This symmetric strategy provides high-quality, annotation-free training data from both perspectives.

Based on the unified triplets $(a, c^+, c^-)$, we optimize the model $f_\theta$ using a multi-task loss function. The first component is a pairwise ranking loss $\mathcal{L}_{\text{pair}}$, which encourages the anchor $a$ to be closer to the positive candidate $c^+$ than to the negative $c^-$ in the embedding space. Formally:
\begin{equation}
\mathcal{L}_{\text{pair}} = -\log \sigma \left( \frac{s(a, c^+) - s(a, c^-)}{\tau} \right),
\end{equation}
where $s(\cdot, \cdot)$ denotes the cosine similarity score and $\tau$ is a temperature hyper-parameter.
The second component is a contrastive cross-entropy loss $\mathcal{L}_{\text{ce}}$, which enhances the model’s discriminative power by pulling the anchor and positive candidate together while pushing away other candidates in the batch. Formally:
\begin{equation} 
\mathcal{L}_{\text{ce}} = -\log \frac{\exp(s(a, c^+)/\tau)}{\sum_{c \in \mathcal{B}} \exp(s(a, c)/\tau)}, 
\end{equation}
where $\mathcal B$ denotes the set of all candidates within the mini-batch.
The final training objective is a weighted sum of the two terms: $\mathcal{L}=\mathcal{L}_{\text{pair}}+\lambda_{\text{ce}}\mathcal{L}_{\text{ce}}$.
To ensure efficient domain adaptation, we implement $f_\theta$ by fine-tuning a pre-trained embedding model via Low-Rank Adaptation~\cite{lora}.

\section{Experiments}

Due to space limitations, we provide the ablation study in Appendix \ref{subsec:ablation_study}, the analysis of pooling strategies in Appendix \ref{subsec:pooling_analysis}, and the evaluation of SPECTER2 variants in Appendix \ref{subsec:specter2_variant_analysis}.

\subsection{Experimental Setup}

\paragraph{Dataset.}
For our method, both the training and validation data are generated from our data construction pipeline, without any human annotation. The test data are obtained from the \bench  and the CMU gold standard dataset. 
From both datasets, we derive pairwise preferences from sparse labels for ranking evaluation as stated in Section~\ref{sec:benchmark}.
To avoid data leakage, we utilize a paper-level holdout protocol: test papers are strictly excluded from all training data construction, while reviewers may overlap across splits, reflecting the practical setting of ranking known reviewers for unseen manuscripts.

\paragraph{Reviewer profiles.}
For each reviewer $r$, the profile text $x_r$ is synthesized from their publications over the preceding two years. In our approach, $x_r$ is a keyword-based profile generated via the pipeline detailed in Section~\ref{sec:profiling}, utilizing Qwen3-Max\footnote{https://help.aliyun.com/zh/model-studio/qwen-api-reference} and GLM4.6\footnote{https://docs.bigmodel.cn/cn/guide/models/text/glm-4.6} as backbone Large Language Models (LLMs). To maintain experimental consistency and ensure a fair comparison, the publication history for all baseline methods is restricted to the same two-year window with ours.

\paragraph{Training details.}
We train Qwen3-Embedding-8B and Qwen3-Embedding-0.6B~\cite{qwen3embedding} with LoRA adapters
on 3,000 dual-view heuristic preference tuples. Detailed hyperparameter settings are provided in Appendix \ref{subsec:Experiment_settings}.

\begin{table*}[t]
\caption{
    \textbf{Main Results.} Comparison of our method against state-of-the-art baselines, where our variants are denoted as [Profiling LLM] + [Embedding Model].
    The \textbf{best results} are bold, and the \underline{runner-up results} are underlined.
    For embedding-based methods, we report results using the pooling strategy that achieves the best results. 
}
  \centering
  \resizebox{0.9\textwidth}{!}{
    \begin{tabular}{l c c c g c c c  g}
      \toprule
      
      & \multicolumn{4}{c}{\textbf{Loss} ($\downarrow$)} & \multicolumn{4}{c}{\textbf{Precision} ($\uparrow$)} \\
      
      \cmidrule(lr){2-5} \cmidrule(lr){6-9}
      
      \textbf{Algorithm} & \textbf{LR-PC} & \textbf{LR-RC} & \textbf{Gold} & \textbf{Avg.} & \textbf{LR-PC} & \textbf{LR-RC} & \textbf{Gold} & \textbf{Avg.} \\
      \midrule
      \rowcolor[HTML]{F2F2F2}
      \multicolumn{9}{l}{\textbf{Statistical-based Methods}} \\
      TPMS & 0.2646 & 0.2333 & 0.2811 & 0.2597 & 70.28\% & 72.30\% & 71.89\% & 71.49\% \\
      \midrule
      \rowcolor[HTML]{F2F2F2}
      \multicolumn{9}{l}{\textbf{Embedding-based Methods}} \\
      ACL & 0.3338 & 0.3038 & 0.3163 & 0.3180 & 61.32\% & 65.96\% & 68.37\% & 65.22\% \\
      CoF & 0.2939 & 0.2218 & 0.2564 & 0.2574 & 65.57\% & 73.31\% & 74.36\% & 71.08\% \\
      BERTScore & 0.2846 & 0.339 & 0.3216 & 0.3153 & 65.57\% & 62.08\% & 67.84\% & 65.16\% \\
      SciBERT   & 0.4016 & 0.4410 & 0.3505 & 0.3977 & 57.55\% & 55.15\% & 64.95\% & 59.22\% \\
      SciNCL    & 0.2354 & 0.2114 & 0.2141 & 0.2203 & 69.34\% & 73.90\% & \underline{78.59}\% & 73.64\% \\
      SPECTER   & 0.2048 & 0.2171 & 0.2672 & 0.2297 & 73.58\% & 73.31\% & 73.28\% & 73.39\% \\
      SPECTER2 PRX & 0.1902 & 0.2176 & 0.2144 & 0.2074 & 74.06\% & 72.89\% & 78.56\% & 75.17\% \\
      \midrule
      \rowcolor[HTML]{F2F2F2}
      \multicolumn{9}{l}{\textbf{LLM-based Methods}} \\
      DeepSeek-V3.2 & 0.2779 & 0.2351 & 0.2237 & 0.2456 & 50.00\% & 53.89\% & 77.36\% & 60.42\% \\
      Qwen3-max     & 0.2713 & 0.2289 & 0.2246 & 0.2416 & 47.17\% & 55.32\% & 77.54\% & 60.01\% \\
      
      \midrule
      \textbf{GLM-4.6 + RATE-0.6B} & 0.2008 & 0.1989 & 0.2350 & 0.2116 & 74.53\% & 74.83\% & 76.50\% & 75.29\% \\
      \textbf{Qwen3-Max + RATE-0.6B} & 0.1955 & 0.2035 & 0.2378 & 0.2123 & 75.47\% & 74.32\% & 76.22\% &75.34\% \\
      \textbf{GLM-4.6 + RATE-8B} & \underline{0.1875} & \textbf{0.1895} & \underline{0.2125} & \underline{0.1965} & \underline{75.94}\% & \textbf{75.51}\% & 78.05\% & \underline{76.78}\% \\
      \textbf{Qwen3-Max + RATE-8B} & \textbf{0.1795} & \underline{0.1926} & \textbf{0.1991} & \textbf{0.1904} & \textbf{76.89}\% & \underline{75.25}\% &\textbf{80.09}\% & \textbf{77.41}\% \\
      \bottomrule
    \end{tabular}
  }
  \label{tab:main_results}
\end{table*}
\subsection{Baseline Methods}
To provide a comprehensive evaluation, we compare our proposed methods against several state-of-the-art baselines, categorized into three groups:

(i) \textbf{Statistical-based Method}. 
We include {TPMS} \cite{TPMS} that calculates relevance using TF-IDF similarity between a reviewer's publication history and the target paper.
(ii) \textbf{Embedding-based Models}. 
These approaches leverage dense embeddings to capture semantic relevance: 
{BERTScore} \cite{bert-score} utilizes contextual embeddings to measure semantic similarity via soft alignment. 
{SciBERT} \cite{beltagy-etal-2019-scibert} is a BERT-based model specifically pre-trained on scientific corpora.
The {SPECTER family}, including SPECTER \cite{cohan-etal-2020-specter}, SciNCL \cite{scincl}, and SPECTER 2 \cite{SciRepEval_dataset}, enhances scientific representations by leveraging citation links and sophisticated sampling strategies.
{CoF} \cite{CoF} is a factor-aware framework that employs instruction tuning and a coarse-to-fine search strategy.
{ACL} \cite{ACL} utilizes contrastive training on non-contiguous abstract segments to identify similarity.
(iii) \textbf{LLMs}. 
We also evaluate DeepSeek-V3.2\cite{deepseekv3.2} and Qwen3-Max\cite{qwen3-max} by prompting them to score reviewer-paper compatibility in a zero-shot setting. 

For completeness, additional details and specific prompts are provided in Appendices~\ref{sec:appendix_baselines} and \ref{subsec:evaluation_prompt}. We tune hyperparameters of all baselines to maximize their performance on our evaluation datasets.

\subsection{Evaluation Protocol and Metrics}
To evaluate our framework, we employ {expertise-aligned loss}, {precision}, and human evaluation to capture both ranking quality and practical utility. 

Following \citet{CMU}, we use a normalized ranking loss $\mathcal L \in [0,1]$ as our primary metric. We unify author-centric and paper-centric perspectives by constructing preference pairs $(x, y)$ where the ground-truth familiarity $\epsilon_x > \epsilon_y$. The loss penalizes misordered predictions by the magnitude of their label difference:
\begin{equation}
\mathcal L = \frac{\sum_{(x,y) \in \mathcal{P}} \mathcal{I}(s_x < s_y) \cdot |\epsilon_x - \epsilon_y|}{\sum_{(x,y) \in \mathcal{P}} |\epsilon_x - \epsilon_y|},
\end{equation}
where $\mathcal{P}$ is the set of all valid pairs, $\mathcal I$ is the indicator function, and  $s$ is the predicted similarity. This metric represents the ratio of the model's error to that of a worst-case adversarial ranker.
In addition to the weighted loss, we report {Precision}, which measures the ratio of pairs where the model correctly predicts the expertise ordering, providing a direct assessment of the model's accuracy.

Further, we conduct a \textbf{human evaluation} to assess the real-world utility of the assignments. We randomly sample 100 papers and task our algorithm and the baselines with retrieving the top-3 candidates from a potential reviewer pool, consistent with the methodology in Section~\ref{sec:recall} (more details are provided in Appendix \ref{subsec:preference_trial}). We invite human experts to perform blind, pair-wise comparisons of these recommendation lists, determining which algorithm provides more qualified matches. We report the \textbf{Win Rate}, defined as the percentage of cases where our algorithm is preferred or judged superior to the baseline.

\subsection{Main Results}

As illustrated in Table~\ref{tab:main_results}, when utilizing Qwen3-Max as the backbone LLM for reviewer profiling and Qwen3-8B-Embedding as the pre-trained embedding model, our approach achieves an average precision of 77.41\% across two datasets, setting a new state-of-the-art (SOTA) performance. Furthermore, our method demonstrates robust generalizability across various backbone profiling LLMs and pre-trained embedding models, consistently maintaining precision levels above 75\%. In contrast, among all evaluated baselines, only SPECTER2 PRX manages to reach the 75\% threshold, further underscoring the superiority and versatility of our proposed framework.

Regarding the simple word-frequency-based TPMS method, we surprisingly observe that despite its algorithmic simplicity, it exhibits remarkable stability, achieving a precision exceeding 70\% on both datasets. Notably, its average precision even outperforms that of modern Large Language Models (LLMs) with vast knowledge bases, suggesting that precise term matching remains a dominant factor in reviewer assignment, potentially outweighing the complex semantic reasoning provided by general-purpose LLMs.

Regarding embedding-based approaches, methods that incorporate scientific citation network information—such as SciNCL and the SPECTER family—outperform pre-trained embedding models that rely solely on semantic content. This finding underscores the pivotal role of citation relationships in reviewer assignment. Specifically, SPECTER2 PRX achieves the highest performance among these, reaching an average precision of 75.17\%. However, it relies heavily on complex aggregation strategies for paper similarity, posing significant challenges for practical deployment.

We further explore the effectiveness of LLMs in this task. Paradoxically, we find that even the most advanced models with superior reasoning capabilities struggle to fully capture the intricacies of reviewer profiling. While their average loss remains relatively moderate at approximately 0.24, their precision performance is underwhelming, hovering around only 60\% ranking as the second and third lowest among all evaluated methods. This discrepancy suggests that while LLMs can effectively distinguish trivial samples, they fail to differentiate between hard samples. 

\begin{table}[t]
\centering
\caption{Human evaluation results on \bench. Win rate indicates the proportion of cases where our method was preferred over the baseline.}
\label{tab:human_eval}
\resizebox{\linewidth}{!}{\begin{tabular}{ccccc} 
\toprule
\multirow{2}{*}{\textbf{Method}} & \multirow{2}{*}{\textbf{Baseline}} & \multicolumn{3}{c}{\textbf{\bench}} \\
\cmidrule(lr){3-5}
 & & \textbf{Win} & \textbf{Lose} & \textbf{Tie} \\
\midrule
\multirow{3}{*}{\textbf{RATE}} & TPMS & 42\% & 8\% &50\% \\
 & SciNCL & 35\% & 17\% & 48\% \\
 & SPECTER2 PRX & 44\% & 12\% & 44\% \\
\bottomrule
\end{tabular}}
\end{table}

We conduct a \textbf{human evaluation} to assess the real-world utility of the assignments. We randomly sample 100 papers and task our algorithm and the baselines with retrieving the top-3 candidates from a potential reviewer pool, consistent with the methodology in \S~3.2 (more details are provided in Appendix \ref{subsec:preference_trial}). We invite human experts to perform blind, pair-wise comparisons of these recommendation lists, determining which algorithm provides more qualified matches. We report the \textbf{Win Rate}, defined as the percentage of cases where our algorithm is judged superior to the baseline.

The results in Table \ref{tab:human_eval} further validate the practical utility of our approach. Due to human effort and resource constraints, we focus this study on TPMS and the two top performing embedding based baselines. Our method achieves higher win rates against all selected baselines, specifically reaching a 42\% win rate against TPMS and 44\% against SPECTER2 PRX. The consistently low lose rates in human trials ranging from 8\% to 17\% underscore that our algorithm provides reliable reviewer recommendations that are well aligned with senior researchers' professional judgment.

\section{Conclusion}
In this paper, we identify the challenges in the reviewer assignment: a lack of high-fidelity, up-to-date evaluation benchmarks, and the misalignment between common training objectives and the goal of reviewer assignment. 
To address the former, we introduced \bench{}, a high-fidelity contemporary benchmark curated from recent CS manuscripts with five-level self-assessed familiarity ratings obtained via outreach emails. 
To address the latter, we introduce RATE, a keyword-based profiling and an annotation-free dual-view preference optimization framework that fine-tunes an embedding model using weak supervision derived from heuristic retrieval signals. 
Experiments showed consistent gains over strong embedding baselines and prior methods, and ablation studies supported the effectiveness of our constructed training data and the benefits of dual-view preference optimization.

\section*{Limitations}
Our method does not explicitly model author-order signals in collaboration-based evidence (e.g., first/last author vs. middle author), which may weaken proxy signals in fields where author order reflects contribution. In addition, the reviewer profile is built from LLM-extracted keywords, which can be noisy or unstable across domains and may propagate errors to ranking. Finally, the approach may be less reliable for cold-start or sparsely published reviewers.

\section*{Ethical Consideration}
Our work involves collecting human feedback via an email survey in which researchers self-report familiarity/expertise ratings for a set of query manuscripts, which have been approved by the Institutional Review Board (IRB). This survey is independent of any real conference or journal review process: the collected ratings are used solely for research evaluation and do not reveal or affect any double-blind reviewing decisions. Participation was voluntary, and respondents could skip questions or stop at any time without any consequences.

Privacy is a primary concern. We contacted potential annotators using email addresses that are publicly available in their published papers. For data release, we will not distribute email addresses or other direct identifiers; instead, we anonymize annotators with random IDs and release only the information necessary for research replication, including manuscript metadata (e.g., title and abstract), the associated ratings, and the list of each annotator’s publications from the past two years used to construct reviewer profiles. Before release, we conduct a privacy audit to detect and remove any direct identifiers or uniquely identifying fields that may appear in the collected data (e.g., email headers, names if present, affiliations), and ensure that only the specified anonymized fields are released. We do not solicit free-form textual responses; any unexpected sensitive or offensive content will be filtered or redacted prior to release. All manuscript/publication metadata are obtained from publicly available records and used in a manner consistent with their original access conditions, and we will respect any third-party restrictions on redistribution.

We note that releasing recent publication lists may still enable re-identification via linkage to public bibliographic records; we therefore avoid releasing any additional identifying attributes (e.g., affiliations) and explicitly inform participants of this residual risk and the intended research-only use. Upon publication, we will distribute the released artifacts with an explicit license and terms of use (research-only; no re-identification), and ensure compliance with any third-party source terms; when redistribution is restricted, we will release only identifiers/links and data-construction scripts instead of redistributing the underlying content. The released benchmark and derived artifacts are intended only for research on reviewer–manuscript matching (e.g., evaluation and reproducibility), and must not be used for operational reviewer selection or other non-research purposes.

Finally, automated reviewer assignment systems may be misused (e.g., to manipulate reviewer selection) or may amplify existing biases (e.g., favoring highly visible institutions or prolific researchers and disadvantaging early-career authors). We view our system strictly as a decision-support tool rather than a replacement for human oversight. Any deployment should incorporate standard safeguards such as conflict-of-interest checks and program-committee review of final assignments.


\bibliography{custom}

@inproceedings{NIPS_dataset,
author = {Mimno, David and McCallum, Andrew},
title = {Expertise modeling for matching papers with reviewers},
year = {2007},
isbn = {9781595936097},
publisher = {Association for Computing Machinery},
address = {New York, NY, USA},
url = {https://doi.org/10.1145/1281192.1281247},
doi = {10.1145/1281192.1281247},
booktitle = {Proceedings of the 13th ACM SIGKDD International Conference on Knowledge Discovery and Data Mining},
pages = {500–509},
numpages = {10},
location = {San Jose, California, USA},
series = {KDD '07}
}

@inproceedings{SciRepEval_dataset,
    title = "{S}ci{R}ep{E}val: A Multi-Format Benchmark for Scientific Document Representations",
    author = "Singh, Amanpreet  and
      D{'}Arcy, Mike  and
      Cohan, Arman  and
      Downey, Doug  and
      Feldman, Sergey",
    editor = "Bouamor, Houda  and
      Pino, Juan  and
      Bali, Kalika",
    booktitle = "Proceedings of the 2023 Conference on Empirical Methods in Natural Language Processing",
    month = dec,
    year = "2023",
    address = "Singapore",
    publisher = "Association for Computational Linguistics",
    url = "https://aclanthology.org/2023.emnlp-main.338/",
    doi = "10.18653/v1/2023.emnlp-main.338",
    pages = "5548--5566",
}

@inproceedings{SIGIR_dataset,
author = {Karimzadehgan, Maryam and Zhai, ChengXiang and Belford, Geneva},
title = {Multi-aspect expertise matching for review assignment},
year = {2008},
isbn = {9781595939913},
publisher = {Association for Computing Machinery},
address = {New York, NY, USA},
url = {https://doi.org/10.1145/1458082.1458230},
doi = {10.1145/1458082.1458230},
booktitle = {Proceedings of the 17th ACM Conference on Information and Knowledge Management},
pages = {1113–1122},
numpages = {10},
location = {Napa Valley, California, USA},
series = {CIKM '08}
}

@inproceedings{CoF,
author = {Zhang, Yu and Shen, Yanzhen and Kang, SeongKu and Chen, Xiusi and Jin, Bowen and Han, Jiawei},
title = {Chain-of-Factors Paper-Reviewer Matching},
year = {2025},
isbn = {9798400712746},
publisher = {Association for Computing Machinery},
address = {New York, NY, USA},
url = {https://doi.org/10.1145/3696410.3714708},
doi = {10.1145/3696410.3714708},
booktitle = {Proceedings of the ACM on Web Conference 2025},
pages = {1901–1910},
numpages = {10},
location = {Sydney NSW, Australia},
series = {WWW '25}
}

@inproceedings{lopez2009grobid,
    title = {{GROBID}: Combining Automatic Bibliographic Data Recognition and Term Extraction for Scholarship Publications},
    author = {Lopez, Patrice},
    booktitle = {Proceedings of the 13th European Conference on Research and Advanced Technology for Digital Libraries (ECDL 2009)},
    year = {2009},
    pages = {473--474}
}

@inproceedings{cohan-etal-2020-specter,
    title = "{SPECTER}: Document-level Representation Learning using Citation-informed Transformers",
    author = "Cohan, Arman  and
      Feldman, Sergey  and
      Beltagy, Iz  and
      Downey, Doug  and
      Weld, Daniel",
    editor = "Jurafsky, Dan  and
      Chai, Joyce  and
      Schluter, Natalie  and
      Tetreault, Joel",
    booktitle = "Proceedings of the 58th Annual Meeting of the Association for Computational Linguistics",
    month = jul,
    year = "2020",
    address = "Online",
    publisher = "Association for Computational Linguistics",
    url = "https://aclanthology.org/2020.acl-main.207/",
    doi = "10.18653/v1/2020.acl-main.207",
    pages = "2270--2282",
}

@misc{CMU,
      title={A Gold Standard Dataset for the Reviewer Assignment Problem}, 
      author={Ivan Stelmakh and John Wieting and Sarina Xi and Graham Neubig and Nihar B. Shah},
      year={2025},
      eprint={2303.16750},
      archivePrefix={arXiv},
      primaryClass={cs.IR},
      url={https://arxiv.org/abs/2303.16750}, 
}

@inproceedings{scincl,
    title = "Neighborhood Contrastive Learning for Scientific Document Representations with Citation Embeddings",
    author = "Ostendorff, Malte  and
      Rethmeier, Nils  and
      Augenstein, Isabelle  and
      Gipp, Bela  and
      Rehm, Georg",
    booktitle = "Proceedings of the 2022 Conference on Empirical Methods in Natural Language Processing",
    year = "2022",
    pages = "11670--11688"

}

@inproceedings{beltagy-etal-2019-scibert,
    title = "{S}ci{BERT}: A Pretrained Language Model for Scientific Text",
    author = "Beltagy, Iz  and
      Lo, Kyle  and
      Cohan, Arman",
    editor = "Inui, Kentaro  and
      Jiang, Jing  and
      Ng, Vincent  and
      Wan, Xiaojun",
    booktitle = "Proceedings of the 2019 Conference on Empirical Methods in Natural Language Processing and the 9th International Joint Conference on Natural Language Processing (EMNLP-IJCNLP)",
    month = nov,
    year = "2019",
    address = "Hong Kong, China",
    publisher = "Association for Computational Linguistics",
    url = "https://aclanthology.org/D19-1371/",
    doi = "10.18653/v1/D19-1371",
    pages = "3615--3620",
}

@inproceedings{10.1145/133160.133205,
author = {Dumais, Susan T. and Nielsen, Jakob},
title = {Automating the assignment of submitted manuscripts to reviewers},
year = {1992},
isbn = {0897915232},
publisher = {Association for Computing Machinery},
address = {New York, NY, USA},
url = {https://doi.org/10.1145/133160.133205},
doi = {10.1145/133160.133205},
booktitle = {Proceedings of the 15th Annual International ACM SIGIR Conference on Research and Development in Information Retrieval},
pages = {233–244},
numpages = {12},
location = {Copenhagen, Denmark},
series = {SIGIR '92}
}

@inproceedings{10.1145/1458082.1458127,
author = {Rodriguez, Marko A. and Bollen, Johan},
title = {An algorithm to determine peer-reviewers},
year = {2008},
isbn = {9781595939913},
publisher = {Association for Computing Machinery},
address = {New York, NY, USA},
url = {https://doi.org/10.1145/1458082.1458127},
doi = {10.1145/1458082.1458127},
booktitle = {Proceedings of the 17th ACM Conference on Information and Knowledge Management},
pages = {319–328},
numpages = {10},
location = {Napa Valley, California, USA},
series = {CIKM '08}
}

@misc{OpenReview,
title = {Paper-reviewer affinity modeling for openreview},
howpublished = {\url{https://github.com/openreview/openreview-expertise}},
note = {Accessed: 2025-12-23},
author = {OpenReview},
year = {2022}
}

@article{https://doi.org/10.1002/int.1055,
author = {Benferhat, Salem and Lang, Jérôme},
title = {Conference paper assignment},
journal = {International Journal of Intelligent Systems},
pages = {1183-1192},
year = {2001}
}

@InProceedings{pmlr-v124-fiez20a,
  title = 	 {A SUPER* Algorithm to Optimize Paper Bidding in Peer Review},
  author =       {Fiez, Tanner and Shah, Nihar and Ratliff, Lillian},
  booktitle = 	 {Proceedings of the 36th Conference on Uncertainty in Artificial Intelligence (UAI)},
  pages = 	 {580--589},
  year = 	 {2020}
}

@article{Tayal2014,
  author  = {Tayal, Devendra Kumar and Saxena, P. C. and Sharma, Ankita and Khanna, Garima and Gupta, Shubhangi},
  title   = {New method for solving reviewer assignment problem using type-2 fuzzy sets and fuzzy functions},
  journal = {Applied Intelligence},
  year    = {2014},
  pages   = {54--73}
}

@article{Tan2021,
  author  = {Tan, Shicheng and Duan, Zhen and Zhao, Shu and Chen, Jie and Zhang, Yanping},
  title   = {Improved reviewer assignment based on both word and semantic features},
  journal = {Information Retrieval Journal},
  year    = {2021},
  pages   = {175--204}
}

@INPROCEEDINGS{6961199,
  author={Xu, Yunhong and Du, Yuanwei},
  booktitle={2013 Sixth International Conference on Business Intelligence and Financial Engineering}, 
  title={A Three-Layer Network Model for Reviewer Recommendation}, 
  year={2013},
  pages={552-556},
  doi={10.1109/BIFE.2013.115}
}

@inproceedings{5d21c3eea4d146f6a345dd044b4ff829,
title = "The AI conference paper assignment problem",
author = "Judy Goldsmith and Sloan, \{Robert H.\}",
year = "2007",
language = "English",
isbn = "9781577353379",
series = "AAAI Workshop - Technical Report",
pages = "53--57",
booktitle = "Preference Handling for Artificial Intelligence - Papers from the 2007 AAAI Workshop, Technical Report",
note = "2007 AAAI Workshop ; Conference date: 22-07-2007 Through 22-07-2007",
}

@inproceedings{TPMS,
  title={The Toronto Paper Matching System: An automated paper-reviewer assignment system},
  author={Laurent Charlin and Richard S. Zemel},
  year={2013},
  url={https://api.semanticscholar.org/CorpusID:680003}
}

@inproceedings{10.1145/1645953.1646207,
author = {Karimzadehgan, Maryam and Zhai, ChengXiang},
title = {Constrained multi-aspect expertise matching for committee review assignment},
year = {2009},
isbn = {9781605585123},
publisher = {Association for Computing Machinery},
address = {New York, NY, USA},
url = {https://doi.org/10.1145/1645953.1646207},
doi = {10.1145/1645953.1646207},
booktitle = {Proceedings of the 18th ACM Conference on Information and Knowledge Management},
pages = {1697–1700},
numpages = {4},
location = {Hong Kong, China},
series = {CIKM '09}
}

@article{Garg2010,
  author  = {Garg, Naveen and Kavitha, Telikepalli and Kumar, Amit and Mehlhorn, Kurt and Mestre, Julián},
  title   = {Assigning Papers to Referees},
  journal = {Algorithmica},
  year    = {2010},
  pages   = {119--136}
}

@article{Stelmakh2018PeerReview4AllFA,
author = {Stelmakh, Ivan and Shah, Nihar and Singh, Aarti},
title = {PeerReview4All: fair and accurate reviewer assignment in peer review},
year = {2021},
issue_date = {January 2021},
publisher = {JMLR.org},
volume = {22},
number = {1},
issn = {1532-4435},
month = jan,
articleno = {163},
numpages = {66},
}

@article{Thurner_2011,
   title={Peer-review in a world with rational scientists: Toward selection of the average},
   volume={84},
   ISSN={1434-6036},
   url={http://dx.doi.org/10.1140/epjb/e2011-20545-7},
   DOI={10.1140/epjb/e2011-20545-7},
   number={4},
   journal={The European Physical Journal B},
   publisher={Springer Science and Business Media LLC},
   author={Thurner, S. and Hanel, R.},
   year={2011},
  pages={707–711} 
}

@article{Black1998,
  author   = {Black, N. and van Rooyen, S. and Godlee, F. and Smith, R. and Evans, S.},
  title    = {What makes a good reviewer and a good review for a general medical journal?},
  journal  = {JAMA},
  year     = {1998},
  pages    = {231--233},
}

@INPROCEEDINGS{10.5555/2888619.2889159,
  author={Bianchi, Federico and Squazzoni, Flaminio},
  booktitle={2015 Winter Simulation Conference (WSC)}, 
  title={Is three better than one? simulating the effect of reviewer selection and behavior on the quality and efficiency of peer review}, 
  year={2015},
  pages={4081-4089},
  doi={10.1109/WSC.2015.7408561}
}

@inproceedings{Shah2022AnOO,
  title={An Overview of Challenges, Experiments, and Computational Solutions in Peer Review (Extended Version)},
  author={Nihar B. Shah},
  year={2022},
  url={https://api.semanticscholar.org/CorpusID:255751797}
}

@article{doi:10.1177/01655515231176668,
author = {Ana Carolina Ribeiro and Amanda Sizo and Luís Paulo Reis},
title ={Investigating the reviewer assignment problem: A systematic literature review},
journal = {Journal of Information Science},
year = {2023},
}

@article{Hsieh2024VulnerabilityOT,
  title={Vulnerability of Text-Matching in ML/AI Conference Reviewer Assignments to Collusions},
  author={Jhih-Yi Hsieh and Aditi Raghunathan and Nihar B. Shah},
  journal={ArXiv},
  year={2024},
}

@inproceedings{bert-score,
  title={BERTScore: Evaluating Text Generation with BERT},
  author={Tianyi Zhang* and Varsha Kishore* and Felix Wu* and Kilian Q. Weinberger and Yoav Artzi},
  booktitle={International Conference on Learning Representations},
  year={2020},
  url={https://openreview.net/forum?id=SkeHuCVFDr}
}

@misc{ACL,
title = {ACL Reviewer Matching},
howpublished = {\url{https://github.com/acl-org/reviewer-paper-matching}},
note = {Accessed: 2025-12-23},
author = {Graham Neubig and John Wieting and Arya McCarthy and Amanda Stent and Natalie Schluter and Trevor Cohn.},
year = {2021}
}

@misc{faiss,
      title={The Faiss library}, 
      author={Matthijs Douze and Alexandr Guzhva and Chengqi Deng and Jeff Johnson and Gergely Szilvasy and Pierre-Emmanuel Mazaré and Maria Lomeli and Lucas Hosseini and Hervé Jégou},
      year={2025},
      eprint={2401.08281},
      archivePrefix={arXiv},
      primaryClass={cs.LG},
      url={https://arxiv.org/abs/2401.08281}, 
}

@inproceedings{Garine2014TheCA,
  title={The comprehensive assessment of Team member effectiveness (catme): personality predicting teamwork competencies},
  author={Armando Garrido Filipe Garine},
  year={2014},
  url={https://api.semanticscholar.org/CorpusID:142239876}
}

@article{https://doi.org/10.1002/ets2.12152,
author = {Kell, Harrison J. and Martin-Raugh, Michelle P. and Carney, Lauren M. and Inglese, Patricia A. and Chen, Lei and Feng, Gary},
title = {Exploring Methods for Developing Behaviorally Anchored Rating Scales for Evaluating Structured Interview Performance},
pages = {1-26},
year = {2017}
}

@inproceedings{dual,
author = {Li, Yizhi and Liu, Zhenghao and Xiong, Chenyan and Liu, Zhiyuan},
title = {More Robust Dense Retrieval with Contrastive Dual Learning},
year = {2021},
isbn = {9781450386111},
publisher = {Association for Computing Machinery},
address = {New York, NY, USA},
url = {https://doi.org/10.1145/3471158.3472245},
doi = {10.1145/3471158.3472245},
booktitle = {Proceedings of the 2021 ACM SIGIR International Conference on Theory of Information Retrieval},
pages = {287–296},
numpages = {10},
location = {Virtual Event, Canada},
series = {ICTIR '21}
}

@article{10.1561/1500000019,
author = {Robertson, Stephen and Zaragoza, Hugo},
title = {The Probabilistic Relevance Framework: BM25 and Beyond},
year = {2009},
issue_date = {April 2009},
publisher = {Now Publishers Inc.},
address = {Hanover, MA, USA},
volume = {3},
number = {4},
issn = {1554-0669},
url = {https://doi.org/10.1561/1500000019},
doi = {10.1561/1500000019},
journal = {Found. Trends Inf. Retr.},
month = apr,
pages = {333–389},
numpages = {57}
}

@misc{fensore2025evaluatinghybridretrievalaugmented,
      title={Evaluating Hybrid Retrieval Augmented Generation using Dynamic Test Sets: LiveRAG Challenge}, 
      author={Chase Fensore and Kaustubh Dhole and Joyce C Ho and Eugene Agichtein},
      year={2025},
      eprint={2506.22644},
      archivePrefix={arXiv},
      primaryClass={cs.CL},
      url={https://arxiv.org/abs/2506.22644}, 
}

@inproceedings{aitymbetov-zorbas-2025-autonomous,
    title = "Autonomous Machine Learning-Based Peer Reviewer Selection System",
    author = "Aitymbetov, Nurmukhammed  and
      Zorbas, Dimitrios",
    editor = "Rambow, Owen  and
      Wanner, Leo  and
      Apidianaki, Marianna  and
      Al-Khalifa, Hend  and
      Eugenio, Barbara Di  and
      Schockaert, Steven  and
      Mather, Brodie  and
      Dras, Mark",
    booktitle = "Proceedings of the 31st International Conference on Computational Linguistics: System Demonstrations",
    month = jan,
    year = "2025",
    address = "Abu Dhabi, UAE",
    publisher = "Association for Computational Linguistics",
    url = "https://aclanthology.org/2025.coling-demos.20/",
    pages = "199--207",
}

@INPROCEEDINGS{5616179,
  author={Tang, Wenbin and Tang, Jie and Tan, Chenhao},
  booktitle={2010 IEEE/WIC/ACM International Conference on Web Intelligence and Intelligent Agent Technology}, 
  title={Expertise Matching via Constraint-Based Optimization}, 
  year={2010},
  volume={1},
  pages={34-41},
  doi={10.1109/WI-IAT.2010.133}}

@inproceedings{10.1145/3183713.3193552,
author = {Wu, Siyuan and U., Leong Hou and Bhowmick, Sourav S. and Gatterbauer, Wolfgang},
title = {PISTIS: A Conflict of Interest Declaration and Detection System for Peer Review Management},
year = {2018},
isbn = {9781450347037},
publisher = {Association for Computing Machinery},
address = {New York, NY, USA},
url = {https://doi.org/10.1145/3183713.3193552},
doi = {10.1145/3183713.3193552},
booktitle = {Proceedings of the 2018 International Conference on Management of Data},
pages = {1713–1716},
numpages = {4},
location = {Houston, TX, USA},
series = {SIGMOD '18}
}

@article{10.1016/j.artint.2024.104119,
author = {Leyton-Brown, Kevin and Mausam and Nandwani, Yatin and Zarkoob, Hedayat and Cameron, Chris and Newman, Neil and Raghu, Dinesh},
title = {Matching papers and reviewers at large conferences},
year = {2024},
issue_date = {Jun 2024},
publisher = {Elsevier Science Publishers Ltd.},
address = {GBR},
volume = {331},
number = {C},
issn = {0004-3702},
url = {https://doi.org/10.1016/j.artint.2024.104119},
doi = {10.1016/j.artint.2024.104119},
journal = {Artif. Intell.},
month = jun,
numpages = {24}
}

@article{ye2024dual,
  title={Dual-Intent-View Contrastive Learning for Knowledge-aware Recommendation},
  author={Ye, Xiaoxin and Zhao, Yan and Zheng, Kai and others},
  journal={Information Sciences},
  volume={667},
  pages={120464},
  year={2024},
  publisher={Elsevier},
  doi={10.1016/j.ins.2024.120464},
  url={https://doi.org/10.1016/j.ins.2024.120464}
}

@inproceedings{
lora,
title={Lo{RA}: Low-Rank Adaptation of Large Language Models},
author={Edward J Hu and Yelong Shen and Phillip Wallis and Zeyuan Allen-Zhu and Yuanzhi Li and Shean Wang and Lu Wang and Weizhu Chen},
booktitle={International Conference on Learning Representations},
year={2022},
url={https://openreview.net/forum?id=nZeVKeeFYf9}
}

@misc{deepseekv3.2,
      title={DeepSeek-V3.2: Pushing the Frontier of Open Large Language Models}, 
      author={DeepSeek-AI and Aixin Liu and Aoxue Mei and Bangcai Lin and Bing Xue and Bingxuan Wang and Bingzheng Xu and Bochao Wu and Bowei Zhang and Chaofan Lin and Chen Dong and Chengda Lu and Chenggang Zhao and Chengqi Deng and Chenhao Xu and Chong Ruan and Damai Dai and Daya Guo and Dejian Yang and Deli Chen and Erhang Li and Fangqi Zhou and Fangyun Lin and Fucong Dai and Guangbo Hao and Guanting Chen and Guowei Li and H. Zhang and Hanwei Xu and Hao Li and Haofen Liang and Haoran Wei and Haowei Zhang and Haowen Luo and Haozhe Ji and Honghui Ding and Hongxuan Tang and Huanqi Cao and Huazuo Gao and Hui Qu and Hui Zeng and Jialiang Huang and Jiashi Li and Jiaxin Xu and Jiewen Hu and Jingchang Chen and Jingting Xiang and Jingyang Yuan and Jingyuan Cheng and Jinhua Zhu and Jun Ran and Junguang Jiang and Junjie Qiu and Junlong Li and Junxiao Song and Kai Dong and Kaige Gao and Kang Guan and Kexin Huang and Kexing Zhou and Kezhao Huang and Kuai Yu and Lean Wang and Lecong Zhang and Lei Wang and Liang Zhao and Liangsheng Yin and Lihua Guo and Lingxiao Luo and Linwang Ma and Litong Wang and Liyue Zhang and M. S. Di and M. Y Xu and Mingchuan Zhang and Minghua Zhang and Minghui Tang and Mingxu Zhou and Panpan Huang and Peixin Cong and Peiyi Wang and Qiancheng Wang and Qihao Zhu and Qingyang Li and Qinyu Chen and Qiushi Du and Ruiling Xu and Ruiqi Ge and Ruisong Zhang and Ruizhe Pan and Runji Wang and Runqiu Yin and Runxin Xu and Ruomeng Shen and Ruoyu Zhang and S. H. Liu and Shanghao Lu and Shangyan Zhou and Shanhuang Chen and Shaofei Cai and Shaoyuan Chen and Shengding Hu and Shengyu Liu and Shiqiang Hu and Shirong Ma and Shiyu Wang and Shuiping Yu and Shunfeng Zhou and Shuting Pan and Songyang Zhou and Tao Ni and Tao Yun and Tian Pei and Tian Ye and Tianyuan Yue and Wangding Zeng and Wen Liu and Wenfeng Liang and Wenjie Pang and Wenjing Luo and Wenjun Gao and Wentao Zhang and Xi Gao and Xiangwen Wang and Xiao Bi and Xiaodong Liu and Xiaohan Wang and Xiaokang Chen and Xiaokang Zhang and Xiaotao Nie and Xin Cheng and Xin Liu and Xin Xie and Xingchao Liu and Xingkai Yu and Xingyou Li and Xinyu Yang and Xinyuan Li and Xu Chen and Xuecheng Su and Xuehai Pan and Xuheng Lin and Xuwei Fu and Y. Q. Wang and Yang Zhang and Yanhong Xu and Yanru Ma and Yao Li and Yao Li and Yao Zhao and Yaofeng Sun and Yaohui Wang and Yi Qian and Yi Yu and Yichao Zhang and Yifan Ding and Yifan Shi and Yiliang Xiong and Ying He and Ying Zhou and Yinmin Zhong and Yishi Piao and Yisong Wang and Yixiao Chen and Yixuan Tan and Yixuan Wei and Yiyang Ma and Yiyuan Liu and Yonglun Yang and Yongqiang Guo and Yongtong Wu and Yu Wu and Yuan Cheng and Yuan Ou and Yuanfan Xu and Yuduan Wang and Yue Gong and Yuhan Wu and Yuheng Zou and Yukun Li and Yunfan Xiong and Yuxiang Luo and Yuxiang You and Yuxuan Liu and Yuyang Zhou and Z. F. Wu and Z. Z. Ren and Zehua Zhao and Zehui Ren and Zhangli Sha and Zhe Fu and Zhean Xu and Zhenda Xie and Zhengyan Zhang and Zhewen Hao and Zhibin Gou and Zhicheng Ma and Zhigang Yan and Zhihong Shao and Zhixian Huang and Zhiyu Wu and Zhuoshu Li and Zhuping Zhang and Zian Xu and Zihao Wang and Zihui Gu and Zijia Zhu and Zilin Li and Zipeng Zhang and Ziwei Xie and Ziyi Gao and Zizheng Pan and Zongqing Yao and Bei Feng and Hui Li and J. L. Cai and Jiaqi Ni and Lei Xu and Meng Li and Ning Tian and R. J. Chen and R. L. Jin and S. S. Li and Shuang Zhou and Tianyu Sun and X. Q. Li and Xiangyue Jin and Xiaojin Shen and Xiaosha Chen and Xinnan Song and Xinyi Zhou and Y. X. Zhu and Yanping Huang and Yaohui Li and Yi Zheng and Yuchen Zhu and Yunxian Ma and Zhen Huang and Zhipeng Xu and Zhongyu Zhang and Dongjie Ji and Jian Liang and Jianzhong Guo and Jin Chen and Leyi Xia and Miaojun Wang and Mingming Li and Peng Zhang and Ruyi Chen and Shangmian Sun and Shaoqing Wu and Shengfeng Ye and T. Wang and W. L. Xiao and Wei An and Xianzu Wang and Xiaowen Sun and Xiaoxiang Wang and Ying Tang and Yukun Zha and Zekai Zhang and Zhe Ju and Zhen Zhang and Zihua Qu},
      year={2025},
      eprint={2512.02556},
      archivePrefix={arXiv},
      primaryClass={cs.CL},
      url={https://arxiv.org/abs/2512.02556}, 
}

@article{qwen3embedding,
  title={Qwen3 Embedding: Advancing Text Embedding and Reranking Through Foundation Models},
  author={Zhang, Yanzhao and Li, Mingxin and Long, Dingkun and Zhang, Xin and Lin, Huan and Yang, Baosong and Xie, Pengjun and Yang, An and Liu, Dayiheng and Lin, Junyang and Huang, Fei and Zhou, Jingren},
  journal={arXiv preprint arXiv:2506.05176},
  year={2025}
}

@misc{qwen3-max,
title = {Qwen3-Max},
howpublished = {\url{https://qwen.ai/apiplatform}},
note = {Accessed: 2025-12-23},
author = {Qwen},
year = {2025}
}

\clearpage
\appendix

\section{Additional Experimental}

\subsection{Ablation Study}
\label{subsec:ablation_study}

To quantify the contribution of our dual-view training strategy and the impact of preference-based fine-tuning, we evaluate the following four configurations:
(1) \textbf{Pretrained (Zero-shot)} uses the original Qwen3-Embedding-8B model without any fine-tuning, serving as a zero-shot baseline to quantify the domain gap;
(2) \textbf{Paper-centric only} fine-tunes the embedding using only paper-centric preference triples $(p,r^+,r^-)$, corresponding to the conventional retrieval-to-reviewer ranking paradigm;
(3) \textbf{Reviewer-centric only} fine-tunes the embedding model using only reviewer-centric preference triples $(r,p^+,p^-)$, emphasizing modeling a reviewer’s historical research trajectory; and
(4) \textbf{Dual-view} is our complete model, trained with both paper-centric and reviewer-centric preferences under the unified objective described in \S~4.2.

Table~\ref{tab:ablation_results} shows a substantial domain gap between the off-the-shelf embedding model and reviewer-assignment preferences: the zero-shot \textbf{Pretrained} baseline performs markedly worse than all fine-tuned variants. Both single-view fine-tuning settings yield substantial gains—\textbf{Paper-centric only} reaches 74.17\% average precision, and \textbf{Reviewer-centric only} reaches 75.26\%—indicating that either perspective provides useful preference supervision for reranking. Importantly, \textbf{Dual-view} achieves the best performance across all evaluation subsets, reaching 77.41\% average precision and 0.1904 average loss. Compared to the best single-view variant, dual-view brings an additional +2.15 precision points, suggesting that integrating both paper-centric and reviewer-centric preferences provides additional, complementary training signal beyond either view alone.

\begin{table*}[t]
\caption{
    \textbf{Ablation Study.} Impact of different training views and data settings on \textbf{Qwen3-Embedding-8B}. LR-PC and LR-RC represent our paper-centric and reviewer-centric benchmarks, while Gold refers to the CMU dataset. The average performance columns are highlighted in gray.
}
\centering
\setlength{\tabcolsep}{8pt} 
\renewcommand{\arraystretch}{1.15}
\small
\begin{tabular}{l c c c g c c c g}
\toprule
& \multicolumn{4}{c}{\textbf{Loss} ($\downarrow$)} & \multicolumn{4}{c}{\textbf{Precision} ($\uparrow$)} \\
\cmidrule(lr){2-5} \cmidrule(lr){6-9}
\textbf{Setting} & \textbf{LR-PC} & \textbf{LR-RC} & \textbf{Gold} & \textbf{Avg.} & \textbf{LR-PC} & \textbf{LR-RC} & \textbf{Gold} & \textbf{Avg.} \\
\midrule
(1) Pretrained (Zero-shot)   & 0.3064  & 0.3705  & 0.4307  & 0.3692 & 64.46\% & 61.73\% & 56.93\%  & 61.04\% \\
(2) Paper-centric Only     & 0.2247 & 0.2114 & 0.2181 & 0.2181  & 70.75\% & \underline{73.56}\% & 78.19\% & 74.17\% \\
(3) Reviewer-centric Only  & \underline{0.1955} & \underline{0.2072} & \underline{0.2141} & \underline{0.2056}  & \underline{74.06}\% & 73.14\% & \underline{78.59}\% & \underline{75.26}\% \\
\midrule
(4) \textbf{Dual-view} & \textbf{0.1795} & \textbf{0.1926} & \textbf{0.1991} & \textbf{0.1904} & \textbf{76.89\%} & \textbf{75.25\%} & \textbf{80.09\%}   & \textbf{77.41\%} \\
\bottomrule
\end{tabular}
\label{tab:ablation_results}
\end{table*}

\subsection{Impact of Pooling Strategies}
\label{subsec:pooling_analysis}

To further investigate how different ways of aggregating reviewer expertise affect matching performance, we compared three pooling strategies: Mean [M], 75th Percentile [75], and Max [X]. Table~\ref{tab:pooling_analysis} presents the detailed results across various embedding-based baselines using Loss and Precision metrics.

\begin{table*}[ht]
\caption{
    \textbf{Impact of Pooling Strategies.} Detailed comparison of Mean [M], 75th Percentile [75], and Max [X] pooling strategies across different embedding-based baselines. 
}
  \centering
  \resizebox{1.0\textwidth}{!}{ 
    \begin{tabular}{l l c c c g c c c g}
      \toprule
      
      & & \multicolumn{4}{c}{\textbf{Loss} ($\downarrow$)} & \multicolumn{4}{c}{\textbf{Precision} ($\uparrow$)} \\
      
      \cmidrule(lr){3-6} \cmidrule(lr){7-10}
      
      \textbf{Model} & \textbf{Pool.} & \textbf{LR-PC} & \textbf{LR-RC} & \textbf{Gold} & \textbf{Avg.} & \textbf{LR-PC} & \textbf{LR-RC} & \textbf{Gold} & \textbf{Avg.} \\
      \midrule
      
      BERTScore & [M]  & 0.2846 & 0.3398 & 0.3216 & 0.3153 & 65.57\% & 62.08\% & 67.84\% & 65.16\% \\
                & [75] & 0.2832 & 0.3330 & 0.3414 & 0.3192 & 66.04\% & 62.84\% & 65.86\% & 64.91\% \\
                & [X]  & 0.3311 & 0.3152 & 0.3033 & 0.3165 & 61.32\% & 64.53\% & 69.67\% & 65.17\% \\
      \midrule
      
      SciBERT   & [M]  & 0.4016 & 0.4410 & 0.3505 & 0.3977 & 57.55\% & 55.15\% & 64.95\% & 59.22\% \\
                & [75] & 0.4668 & 0.4515 & 0.3630 & 0.4271 & 52.36\% & 54.05\% & 63.70\% & 56.70\% \\
                & [X]  & 0.4282 & 0.4546 & 0.3449 & 0.4092 & 55.66\% & 53.63\% & 65.51\% & 58.27\% \\
      \midrule
      
      SciNCL    & [M]  & 0.2168 & \underline{0.2077} & 0.2601 & 0.2282 & 71.70\% & 73.99\% & 73.99\% & 73.23\% \\
                & [75] & 0.2061 & \textbf{0.1994} & 0.2663 & 0.2239 & 73.58\% & \textbf{74.49}\% & 73.37\% & 73.81\% \\
                & [X]  & 0.2354 & 0.2114 & \textbf{0.2141} & 0.2203 & 69.34\% & \underline{73.90}\% & \textbf{78.59}\% & 73.94\% \\
      \midrule
      
      SPECTER   & [M]  & 0.2380 & 0.2505 & 0.3115 & 0.2667 & 70.28\% & 70.35\% & 68.85\% & 69.83\% \\
                & [75] & 0.2247 & 0.2364 & 0.2851 & 0.2487 & 72.64\% & 71.96\% & 71.49\% & 72.03\% \\
                & [X]  & 0.2048 & 0.2171 & 0.2672 & 0.2297 & \underline{73.58}\% & 73.31\% & 73.28\% & 73.39\% \\
      \midrule
      
      SPECTER2  & [M]  & 0.2141 & 0.2354 & 0.2507 & 0.2334 & 71.23\% & 71.88\% & 74.93\% & 72.68\% \\
      (PRX)     & [75] & \underline{0.1981} & 0.2166 & 0.2436 & \underline{0.2194} & \underline{73.58}\% & 73.06\% & 75.64\% & \underline{74.09}\% \\
                & [X]  & \textbf{0.1902} & 0.2176 & \underline{0.2144} & \textbf{0.2074} & \textbf{74.06}\% & 72.89\% & \underline{78.56}\% & \textbf{75.17}\% \\
      
      \bottomrule
    \end{tabular}
  }
  \label{tab:pooling_analysis}
\end{table*}

\subsection{Comprehensive Evaluation of SPECTER2 Variants}
\label{subsec:specter2_variant_analysis}

We evaluated five SPECTER2 adapter variants: Base, Adhoc Query, Classification, Proximity, and Regression. Each variant was tested across the three pooling strategies, with the full performance metrics presented in Table~\ref{tab:specter2_variant_analysis}

\begin{table*}[ht]
\caption{
    \textbf{Comprehensive Evaluation of SPECTER2 Variants.} Comparison of SPECTER2 with different adapters (Base, Adhoc Query, Classification, Proximity, and Regression) under Mean [M], 75th Percentile [75], and Max [X] pooling strategies.
}
  \centering
  \resizebox{1.0\textwidth}{!}{ 

    \begin{tabular}{l l c c c >{\columncolor{gray!15}}c c c c >{\columncolor{gray!15}}c}
    
      \toprule
      
      & & \multicolumn{4}{c}{\textbf{Loss} ($\downarrow$)} & \multicolumn{4}{c}{\textbf{Precision} ($\uparrow$)} \\
      
      \cmidrule(lr){3-6} \cmidrule(lr){7-10}
      
      \textbf{Adapter} & \textbf{Pool.} & \textbf{LR-PC} & \textbf{LR-RC} & \textbf{Gold} & \textbf{Avg.} & \textbf{LR-PC} & \textbf{LR-RC} & \textbf{Gold} & \textbf{Avg.} \\
      \midrule
      
      SPECTER2 (Base) & [M]  & 0.2035 & 0.2255 & 0.2594 & 0.2295 & 72.64\% & 72.72\% & 74.06\% & 73.14\% \\
                      & [75] & 0.2114 & \textbf{0.2082} & 0.2484 & 0.2227 & 71.70\% & \textbf{73.65}\% & 75.16\% & 73.50\% \\
                      & [X]  & 0.2074 & 0.2150 & \underline{0.2307} & \underline{0.2177} & 72.17\% &\underline{73.48}\% & \underline{76.93}\% & \underline{74.19}\% \\
      \midrule
      
      Adhoc Query     & [M]  & 0.2434 & 0.2599 & 0.2817 & 0.2617 & 69.34\% & 69.34\% & 71.83\% & 70.17\% \\
                      & [75] & 0.2434 & 0.2390 & 0.3042 & 0.2622 & 68.87\% & 71.28\% & 69.58\% & 69.91\% \\
                      & [X]  & 0.2646 & 0.2427 & 0.2717 & 0.2597 & 66.04\% & 70.35\% & 72.83\% & 69.74\% \\
      \midrule
      
      Classification  & [M]  & 0.3045 & 0.2677 & 0.2709 & 0.2810 & 63.68\% & 68.33\% & 72.91\% & 68.31\% \\
                      & [75] & 0.2673 & 0.2469 & 0.2639 & 0.2594 & 67.45\% & 69.85\% & 73.61\% & 70.30\% \\
                      & [X]  & 0.2527 & 0.2589 &  0.2893 & 0.2670 & 68.40\% & 69.51\% & 71.07\% & 69.66\% \\
      \midrule

      Proximity (PRX) & [M]  &  0.2141 & 0.2354 & 0.2507 & 0.2334 & 71.23\% & 71.88\% & 74.93\% & 72.68\% \\
                      & [75] & \underline{0.1981} & \underline{0.2166} & 0.2436 & 0.2194 & \underline{73.58}\% & 73.06\% & 75.64\% & 74.09\% \\
                      & [X]  & \textbf{0.1902} & 0.2176 & \textbf{0.2144} & \textbf{0.2074} & \textbf{74.06}\% & 72.89\% & \textbf{78.56}\% & \textbf{75.17}\% \\
      \midrule

      Regression      & [M]  & 0.3697 & 0.3946 & 0.3751 & 0.3798 & 57.08\% & 59.88\% & 62.49\% & 59.82\% \\
                      & [75] & 0.3152 & 0.3559 & 0.3581 & 0.3431 & 59.91\% & 63.09\% & 64.19\% & 62.40\% \\
                      & [X]  & 0.3298 & 0.3591 & 0.3465  & 0.3451 & 59.43\% & 62.50\% & 65.35\% & 62.43\% \\

      \bottomrule
    \end{tabular}
  }
  \label{tab:specter2_variant_analysis}
\end{table*}

\section{Expert Survey Email Template}
\label{sec:appendix_email}

To collect expert feedback for our dataset construction, we sent out standardized inquiry emails to candidates in the reviewer pool.  The template used for this communication, which ensures transparency regarding the research purpose and data usage, is presented in Table~\ref{box:email_template}.

\begin{table}[ht]
\centering
\begin{minipage}{0.98\linewidth}
\centering
\begin{mdframed}[linewidth=0.8pt, innerleftmargin=10pt, innerrightmargin=10pt, innertopmargin=10pt, innerbottommargin=10pt, backgroundcolor=gray!2]
\small
Dear [Scholar Name],

\vspace{0.5em}
We hope this email finds you well. We are a research team from the [Department/Lab Name] at [University/Institute Name].

\vspace{0.5em}
With rapidly growing submissions to AI-related conferences, reviewer assignment has become increasingly challenging. To support research in this area, we are constructing a publicly available dataset to advance automated reviewer assignment systems.

\vspace{0.5em}
Based on your published works (e.g., ``\textit{[Example Publication Title]} ''), our algorithm identified you as a potential expert for the papers listed below.

\vspace{0.5em}
\textbf{Note:} this is NOT a request to review a paper. We only seek to collect expert feedback to help construct the dataset, and the collected data will be used solely for academic research purposes. We would be very grateful if you could click the value in the circle below that best reflects your familiarity with each topic.

\vspace{0.5em}
It takes less than 10 seconds per paper, and your feedback would be incredibly valuable to our study. Thank you very much for your time and help.

\vspace{0.8em}
 Best regards,

\vspace{0.3em}
[Department/Lab Name]

\vspace{0.4em}
[University/Institute Name]
\end{mdframed}
\captionof{table}{Standardized email template sent to scholars for expertise verification.}
\label{box:email_template}
\end{minipage}
\end{table}

\section{Human Evaluation}

\subsection{Behaviorally Anchored Rating Scale}
\label{subsec:expert_scale}

 To ensure a faithful ground truth beyond the limitations of administrative records, we collected expert self-assessments via email using a five-level Behaviorally Anchored Rating Scale (BARS). The specific criteria provided to the participants are defined as follows:  

\begin{itemize}[leftmargin=*, itemsep=0pt]
 \item \textbf{5 - Top Expert:} \textit{I am an active researcher in this sub-field; I have recently published work highly relevant to this paper, or I could write a similar paper myself.}
 \item \textbf{4 - Expert:} \textit{I am very familiar with this field; I could reproduce the method in the paper and accurately judge the quality of its technical details.}
 \item \textbf{3 - Knowledgeable:} \textit{I work or research in a related field; I understand the core concepts but have not published papers or worked on projects in this specific sub-direction.}
 \item \textbf{2 - Vague Familiarity:} \textit{I have heard of this field and can understand the abstract, but I am unfamiliar with the specific methodologies or technical details.}
 \item \textbf{1 - No Expertise:} \textit{I completely do not understand this field and cannot understand the terminology or core logic in the text.}
\end{itemize}

\subsection{Human Preference Trial and Evaluation Details}
\label{subsec:preference_trial}

We conducted a human preference study using pairwise comparisons to evaluate our algorithm against competitive baselines, reporting the results in terms of the \textbf{win rate}.

The evaluation team consisted of five Master's and PhD students with relevant domain expertise and English proficiency. To ensure ethical labor practices, all judges were compensated at a rate exceeding the local minimum wage. Before the study, we briefed them on the task and ensured they fully understood the objectives and provided informed consent. \textbf{To further align judgment standards, each judge was provided with several high-quality evaluation examples as references, demonstrating the application of our criteria  (criteria detailed below)}.

Given the complexity of assessing research expertise, we provided the judges with specific guidelines to ensure consistency. When presented with a query paper and two sets of Top-3 recommended reviewers (alongside their publication histories), judges were instructed to select the set that better satisfied the following criteria:

\begin{itemize}[leftmargin=*, noitemsep]
    \item \textbf{Topic Alignment:} The degree of fit between the reviewers' research backgrounds and the query paper’s core domain, keywords, and technical methodologies.
    \item \textbf{Expertise Depth:} Whether the reviewers have published high-quality work in relevant fields, ensuring they can evaluate technical contributions rather than just surface-level concepts.
    \item \textbf{Complementary Coverage:} The extent to which the Top-3 set collectively covers different facets of the paper (e.g., for a paper on ``RL in Healthcare,'' a mix of RL and medical informatics experts is preferred over three experts in only one area).
\end{itemize}

In cases where the two sets were equally qualified, they were instructed to report a \textbf{Tie}.

\section{Experiment Settings}
\label{subsec:Experiment_settings}

\subsection{Computing Facilities.}
\label{subsec:computing_facilities}
 All experiments are conducted on a single NVIDIA A800-80G GPU. 

\subsection{Hyperparameter Settings}
\label{subsec:hyperparameters}

We fine-tune the Qwen3-Embedding-8B model using LoRA. The key hyperparameters are summarized in Table~\ref{tab:hyperparameters}. The task prompts used for query ($Q$) and reviewer ($R$) retrieval are detailed in the following paragraph.

\begin{table}[ht]
\centering
\footnotesize
\caption{Hyperparameter settings for fine-tuning.}
\begin{tabularx}{\columnwidth}{lX} 
\toprule
\textbf{Category} & \textbf{Hyperparameters (Value)} \\
\midrule
\textbf{LoRA} & $r$: 16, $\alpha$: 32, Dropout: 0.1 \\
\midrule
\textbf{Optimization} & LR: 2.3e-05, Warmup: 0.05, Epochs: 15 \\
                      & Batch: 4, Accumulation: 1, Seed: 622 \\
                      & $\tau$: 0.0634, Patience: 6 \\
\midrule
\textbf{Input}        & Max Len ($Q/R$): 2048, Keywords: 512 \\
                      & Weights: CE (0.915), Pair (1.0) \\
\bottomrule
\end{tabularx}
\label{tab:hyperparameters}
\end{table}

\noindent \textbf{Task Prompts.} For the retrieval task, we use: 
(1) \textit{Query:} ``Given a submission title and abstract, retrieve reviewers whose expertise profile matches and who are familiar with the work.'' 
(2) \textit{Reviewer-Centric:} ``Given a reviewer profile, retrieve papers that match the reviewer's expertise.''

\section{Baselines Methods.}
\label{sec:appendix_baselines}

To ensure the reproducibility of our experiments, we summarize the specific implementation sources and model checkpoints for all baseline methods in Table~\ref{tab:baseline_resources}. 

While most embedding-based models in our study are evaluated using general pooling strategies (i.e., Mean, Max, or Percentile) to aggregate paper-to-paper similarities, we adhere to the original aggregation rules for certain established baselines to ensure a fair comparison:

\begin{itemize}[leftmargin=2em, itemsep=4pt, label=$\circ$]
    \item \textbf{ACL Algorithm:} We follow the established calculation by identifying the top 3 most similar papers and summing their scores weighted by $1/n$, where $n$ is the rank of the paper in terms of similarity.
    \item \textbf{CoF:} This method aggregates multiple relevance components, including semantic, topical, and citation-based features, to calculate the final expertise fit.
\end{itemize}

\clearpage

\begin{table}[ht]
\centering
\footnotesize
\caption{Implementation sources and model checkpoints for baselines.}
\label{tab:baseline_resources}
\begin{tabularx}{\columnwidth}{lX}
\toprule
\textbf{Method} & \textbf{Source / Checkpoint} \\
\midrule
TPMS & \href{https://github.com/niharshah/goldstandard-reviewer-paper-match}{\texttt{niharshah/goldstandard}} \\
SciBERT & \href{https://huggingface.co/allenai/scibert_scivocab_uncased}{\texttt{allenai/scibert\_uncased}} \\
BERTScore & \href{https://github.com/Tiiiger/bert_score}{\texttt{Tiiiger/bert\_score}} \\
SPECTER & \href{https://huggingface.co/allenai/specter}{\texttt{allenai/specter}} \\
SciNCL & \href{https://huggingface.co/malteos/scincl}{\texttt{malteos/scincl}} \\
SPECTER 2.0 & \href{https://huggingface.co/allenai/specter2}{\texttt{allenai/specter2}} \\
ACL Algo. & \href{https://github.com/acl-org/reviewer-paper-matching}{\texttt{acl-org/matching}} \\
CoF & \href{https://github.com/yuzhimanhua/CoF}{\texttt{yuzhimanhua/CoF}} \\
\bottomrule
\end{tabularx}
\end{table}

\section{Large Language Model Prompts}

\subsection{Large Language Model Prompt for Author Disambiguation}
\label{subsec:disambiguation_prompt}

To handle potential author ambiguity and ensure data quality, we utilize an LLM-based clustering approach for author disambiguation. The specific instructions and matching rules used for this task are presented in Table~\ref{box:disambiguation_prompt}.

\begin{figure}[ht]
\centering
\begin{minipage}{0.98\linewidth}
\centering
\begin{mdframed}[linewidth=0.8pt, innerleftmargin=10pt, innerrightmargin=10pt, innertopmargin=10pt, innerbottommargin=10pt, backgroundcolor=gray!5] 
\small
\textbf{Task:} Cluster author instances based on institutional similarity.

\vspace{0.5em}
\textbf{[Clustering Rules]}
\begin{enumerate}[leftmargin=*, itemsep=0pt]
    \item \textbf{Primary Rule:} Focus on \textit{Institutional Similarity}. Instances sharing the same main institution (university, company, research institute) are considered the same author.
    \item \textbf{Institution Matching:} Group name variations (e.g., ``MIT'' = ``Massachusetts Institute of Technology'') and different location formats (e.g., ``Alibaba Group'' = ``Alibaba Inc.'').
    \item \textbf{Department vs. Institution:} Prioritize the main institution over specific departments (e.g., ``Dept. of CS, Stanford'' = ``Stanford'').
    \item \textbf{Empty Institutions:} Instances with empty or null affiliation fields must be placed in separate, individual clusters and never grouped with non-empty ones.
\end{enumerate}

\vspace{0.5em}
\textbf{[Input Data]} \\
\texttt{Instances:} \{instances\}

\vspace{0.8em}
\textbf{[Output Requirements]} \\
Return the results \textbf{strictly} in JSON format:
\begin{quote}
\texttt{\{ ``clusters'': [[1, 2, 3], [4]] \}}
\end{quote}
\textit{Ensure all instance IDs are included. Output only the JSON without any explanatory text.}
\end{mdframed}
\captionof{table}{Prompting template for LLM-based author disambiguation.}
\label{box:disambiguation_prompt}
\end{minipage}
\end{figure}

\subsection{Large Language Model Prompt for Keyword Extraction}
\label{subsec:keyword_extraction_prompt}

To construct comprehensive author profiles for downstream matching, we employ an LLM to extract domain-specific keywords from paper titles and abstracts. This process ensures that the response focus is captured through precise and representative topics. The prompt template is detailed in Table~\ref{box:keyword_prompt}.

\begin{figure}[ht]
\centering
\begin{minipage}{0.98\linewidth}
\centering
\begin{mdframed}[linewidth=0.8pt, innerleftmargin=10pt, innerrightmargin=10pt, innertopmargin=10pt, innerbottommargin=10pt, backgroundcolor=gray!5] 
\small
\textbf{Role:} You are an academic keyword extraction assistant.

\vspace{0.5em}
\textbf{Task:} Extract precise, domain-representative keywords from the research paper's title and abstract. These keywords should represent research domains, subfields, or areas of study most relevant to the paper's focus.

\vspace{0.5em}
\textbf{[Input Data]} \\
\texttt{Title:} \{paper\_title\} \\
\texttt{Abstract:} \{paper\_abstract\}

\vspace{0.8em}
\textbf{[Output Requirements]} \\
Please extract $N$ concise and specific keywords. You must output \textbf{ONLY} the keywords as a comma-separated list, with no additional text, explanations, or formatting.

\vspace{0.5em}
\textit{Example Output: Large Language Models, Information Retrieval, Graph Neural Networks}
\end{mdframed}
\captionof{table}{Prompting template for LLM-based academic keyword extraction.}
\label{box:keyword_prompt}
\end{minipage}
\end{figure}

\subsection{Large Language Model Prompt for Evaluation}
\label{subsec:evaluation_prompt}
For evaluating LLMs in a zero-shot setting, we use a structured prompt to guide the model in scoring reviewer expertise. The template is shown in Table~\ref{box:prompt}.

\begin{figure}[ht]
\centering
\begin{minipage}{0.98\linewidth}
\centering
\begin{mdframed}[linewidth=0.8pt, innerleftmargin=10pt, innerrightmargin=10pt, innertopmargin=10pt, innerbottommargin=10pt, backgroundcolor=gray!5] 
\small
\textbf{Task:} Score the reviewer expertise and fit for reviewing the target paper.

\vspace{0.5em}
\textbf{[Target Paper]} \\
\texttt{Title:} \{paper\_title\} \\
\texttt{Abstract:} \{paper\_abstract\}

\vspace{0.5em}
\textbf{[Reviewer Profile]} \\
\texttt{Name:} \{reviewer\_name\} \\
\texttt{Recent Publications:} \{reviewer\_papers\}

\vspace{0.8em}
\textbf{[Scoring Rubric]} \\
Please choose one integer score from 1 to 5: \\
\textbf{5 (Top Expert):} Active researcher in this specific sub-field; published highly relevant work; capable of writing a similar paper. \\
\textbf{4 (Expert):} Very familiar with the general field; can reproduce methods and judge technical quality. \\
\textbf{3 (Knowledgeable):} Works in a related field; understands core concepts but no direct work in this sub-direction. \\
\textbf{2 (Vague Familiarity):} General awareness; can understand the abstract but unfamiliar with technical nuances. \\
\textbf{1 (No Expertise):} No background; cannot understand terminology or core logic.

\vspace{0.8em}
\textit{At the end of your response, you MUST output only one integer from 1 to 5, and nothing else.}
\end{mdframed}
\captionof{table}{Zero-shot prompting template for LLM-based evaluation.}
\label{box:prompt}
\end{minipage}
\end{figure}

\end{document}